\documentclass[sigconf,nonacm]{acmart}

\usepackage{graphicx}
\usepackage{pifont}
\usepackage{colortbl}
\usepackage{multirow}
\usepackage{makecell}
\usepackage{bm}

\usepackage{algorithm}
\usepackage{algorithmicx}
\usepackage{algpseudocode}

\usepackage[most]{tcolorbox}
\usepackage{bm}
\usepackage{cleveref}

\definecolor{vividcyan}{RGB}{0, 176, 240}
\definecolor{MyPurple}{HTML}{8E44AD}

\newtcolorbox{promptbox}[2][]{
  breakable,
  width=\textwidth,
  colback=black!5,
  colframe=black!60,
  arc=2mm,
  boxrule=0.8pt,
  fonttitle=\bfseries,
  title={#2},
  #1
}

\ccsdesc[500]{Computing methodologies~Computer vision tasks}
\ccsdesc[500]{Computing methodologies~Image manipulation}
\ccsdesc[500]{Security and privacy~Digital forensics}

\keywords{Text-Centric Forgery, Multimodal Reasoning, Explainable AI, MLLM, AI Document Safety}

\title{DocShield: Towards AI Document Safety via Evidence-Grounded Agentic Reasoning}

\author{
    Fanwei Zeng$^1$, 
    Changtao Miao$^1$, 
    Jing Huang$^1$, 
    Zhiya Tan$^{1,2}$, 
    Shutao Gong$^1$, 
    Xiaoming Yu$^1$, 
    Yang Wang$^1$, 
    Weibin Yao$^1$, 
    Joey Tianyi Zhou$^3$, 
    Jianshu Li$^1$, 
    Ying Yan$^1$
}

\email{fanwei.zfw@antgroup.com, jianshu.l@antgroup.com}

\affiliation{
  \institution{
    \parbox{\textwidth}{\centering
      \textsuperscript{1}Ant Group \quad \textsuperscript{2}Nanyang Technological University \\
      \textsuperscript{3}CFAR and IHPC, Agency for Science, Technology and Research (A*STAR), Singapore
    }
  }
  \country{} 
}

\renewcommand\footnotetextcopyrightpermission[1]{} 
\renewcommand{\shortauthors}{Zeng et al.} 

\begin{abstract}
The rapid progress of generative AI has enabled increasingly realistic text-centric image forgeries, posing major challenges to document safety. 
Existing forensic methods mainly rely on visual cues and lack evidence-based reasoning to reveal subtle text manipulations. 
Detection, localization, and explanation are often treated as isolated tasks, limiting reliability and interpretability. 
To tackle these challenges, we propose \textbf{DocShield}, the first unified framework formulating text-centric forgery analysis as a visual–logical co–reasoning problem. 
At its core, a novel \textbf{Cross-Cues-aware Chain of Thought (CCT)} mechanism enables implicit agentic reasoning, iteratively cross-validating visual anomalies with textual semantics to produce consistent, evidence-grounded forensic analysis. 
We further introduce a \textbf{Weighted Multi-Task Reward} for GRPO-based optimization, aligning reasoning structure, spatial evidence, and authenticity prediction. 
Complementing the framework, we construct \textbf{RealText-V1}, a multilingual dataset of document-like text images with pixel-level manipulation masks and expert-level textual explanations. 
Extensive experiments show \textbf{DocShield} significantly outperforms existing methods, improving macro-average F1 by \textbf{41.4\%} over specialized frameworks and \textbf{23.4\%} over GPT-4o on T-IC13, with consistent gains on the challenging T-SROIE benchmark. 
Our dataset, model, and code will be publicly released.
\end{abstract}

\begin{document}
\maketitle

\section{Introduction}
\label{sec:intro}

\begin{figure}[h!]
  \centering
  \includegraphics[width=\columnwidth]{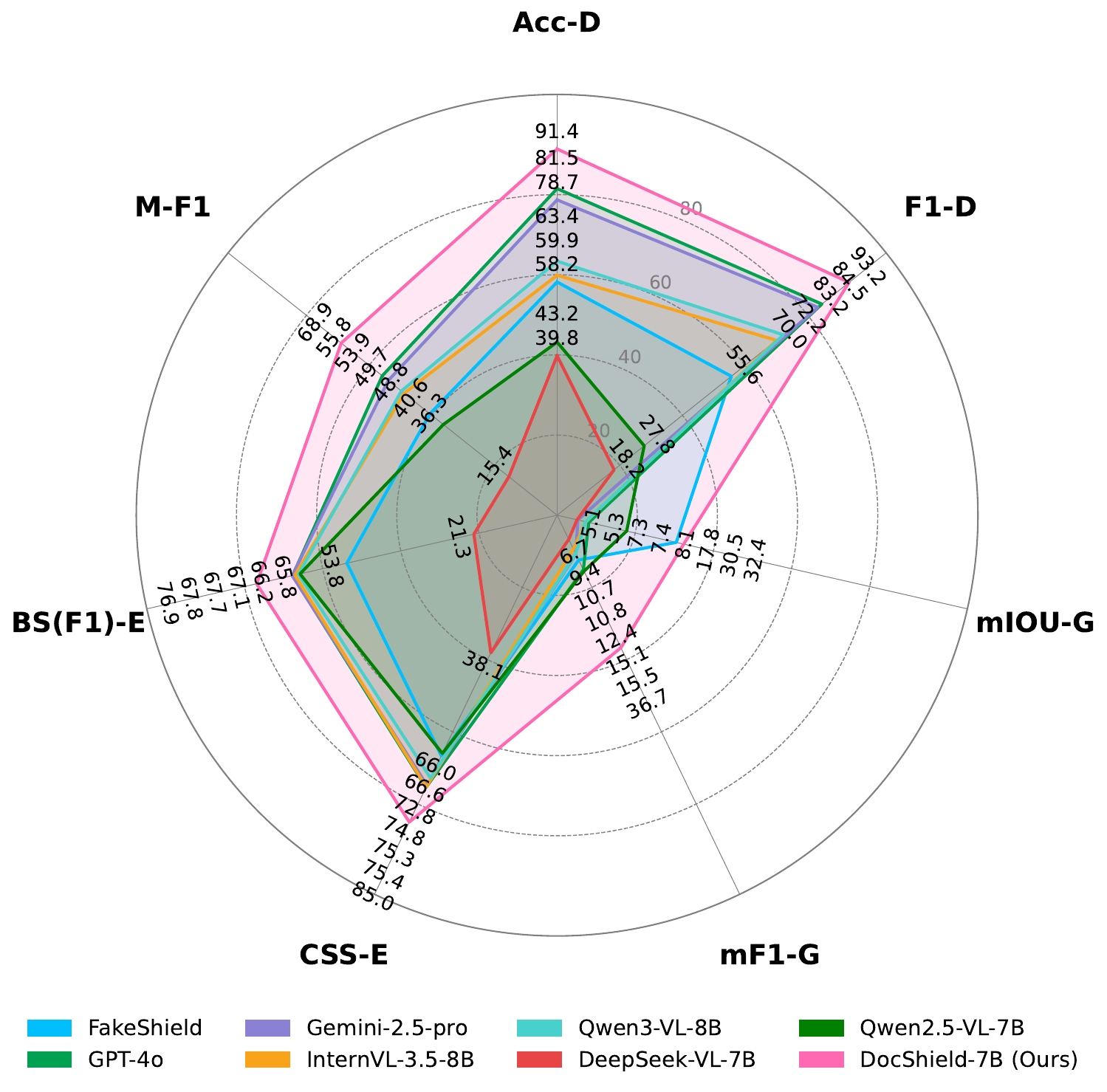}
  \caption{\textbf{Performance comparison on the RealText-V1 benchmark.} DocShield (pink area) achieves state-of-the-art results across Detection (D), Grounding (G), and Explanation (E) tasks. M-F1 (macro-average F1 over the three tasks) serves as the unified evaluation metric, demonstrating DocShield's superior and balanced multi-task capability.}
  \Description{A radar chart comparing DocShield (pink area) against multiple MLLM baselines on RealText benchmark, showing metrics for detection, grounding, and explanation tasks, with DocShield demonstrating superior performance across all dimensions.}
\label{fig:intro_radar_chart}
\end{figure}

The rapid evolution of generative AI, driven by advanced diffusion models~\cite{dalle3, midjourney} and video generation systems~\cite{liu2024sora, ramachandran2024sora}, has significantly lowered the barrier to realistic content manipulation. As a result, \textbf{text-centric image forgeries} have emerged as a critical threat to information authenticity. Unlike generic image manipulation, these attacks target high-stakes, document-like media—such as financial receipts, legal contracts, and official notices—where subtle textual edits can fundamentally alter semantics. This growing risk highlights an urgent need to move \textit{towards AI document safety}: developing forensic systems that can reliably verify both the visual integrity and semantic consistency of document-centric content.

\begin{figure*}[t]
  \centering
  \includegraphics[width=\textwidth, height=0.62\textwidth]{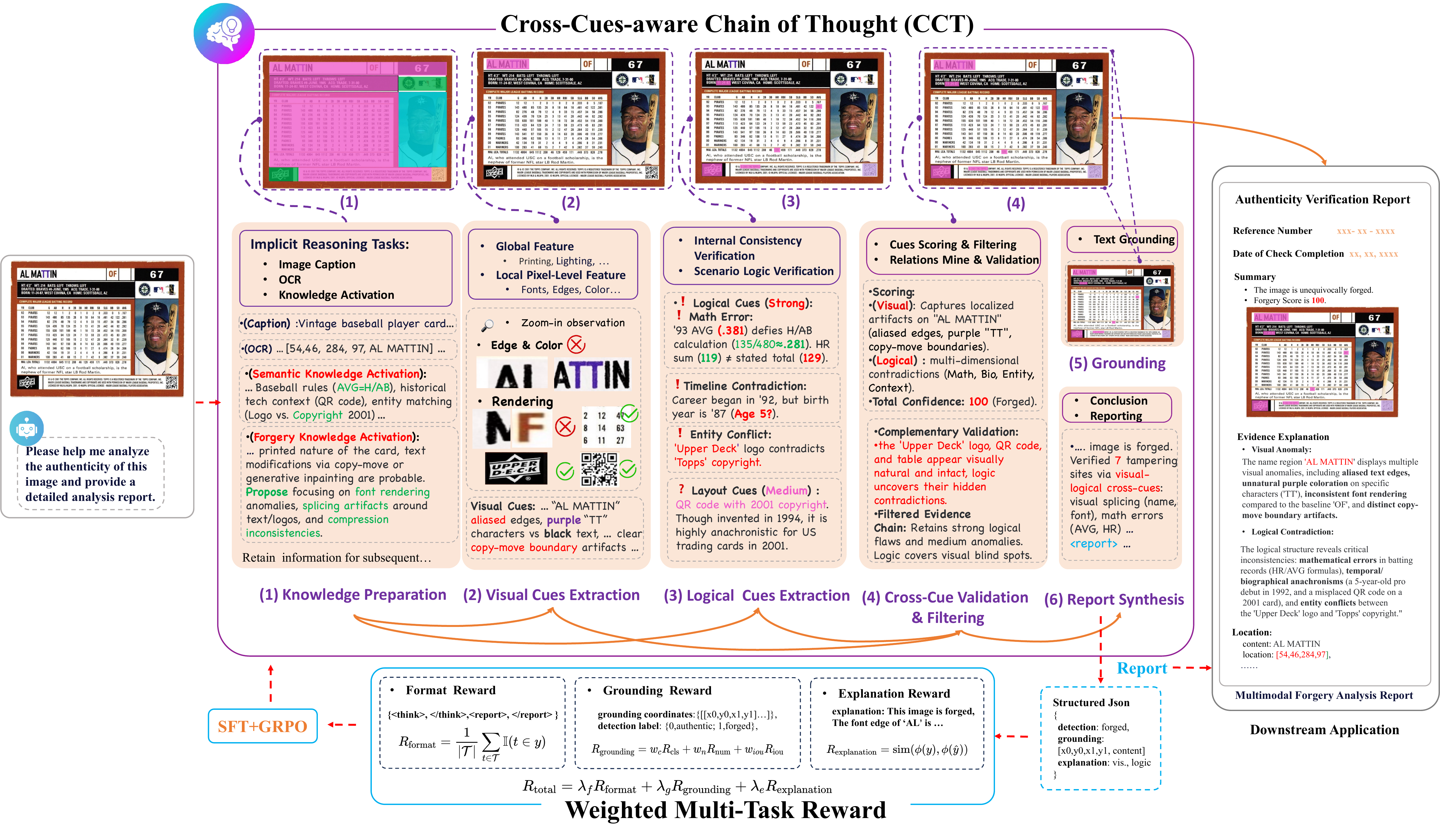}
  \caption{\textbf{Overview of the DocShield framework.} Given an input image and a prompt, the model autoregressively generates a structured, machine-readable \textcolor{vividcyan}{\textit{\textbf{Report}}}. The analytical core is the \textbf{Cross-Cues-aware Chain of Thought (CCT)}, a six-stage mechanism that iteratively extracts and cross-validates visual and logical anomalies. To ensure forensic faithfulness, the framework is optimized via Group Relative Policy Optimization (GRPO) using a \textbf{Weighted Multi-Task Reward}, strictly aligning spatial evidence, format compliance, and explanation fidelity.}
  \Description{Diagram of the DocShield framework showing image input and prompt processed through the Cross-Cues-aware Chain of Thought (CCT) mechanism, including stages for knowledge preparation, visual and logical cue extraction, cross-validation, grounding, and report synthesis. Weighted multi-task reward optimizes the system via GRPO. The structured report can be used for downstream forensic applications.}
\label{fig:framework_overview}
\end{figure*}

Traditional forensic methods typically formulate forgery detection as a coarse-grained binary classification problem, which inherently lacks interpretability~\cite{lampert2006printing, chen2024single, tan2024frequency, wu2019mantra, cozzolino2020noiseprint}. Recent advances in Multimodal Large Language Models (MLLMs) have enabled more explainable analysis~\cite{qu2024textsleuth, xu2025fakeshield, gao2025fakereasoning, kang2026legion}, yet existing approaches suffer from two fundamental limitations. 
\textbf{First, they typically treat detection, spatial grounding, and explanation as discrete, decoupled sub-tasks.} This separation prevents joint optimization and explicit consistency modeling, inevitably causing error propagation and hindering holistic performance.
\textbf{Second, these approaches mostly rely on visual signals
and largely neglect logical contradictions embedded within the
textual content itself.} 
When visual traces are subtle or actively obfuscated by attackers, these models frequently suffer from \textit{reasoning hallucinations}, generating plausible but factually ungrounded explanations (i.e., blind guessing).
As regulatory demands for trustworthy AI continue to grow~\cite{nist2023ai, a&o2024euaiact}, effective document safety requires 
a unified, evidence-grounded paradigm that provides precise localization of manipulations supported by a logical, cross-verified analysis of the clues, as illustrated in Fig.~\ref{fig:framework_overview}.

To overcome these limitations, we propose \textbf{DocShield}, a unified generative framework that \textbf{reformulates text-centric forgery analysis as a visual–logical co-reasoning problem}, marking a step towards principled AI document safety. 
The deep reasoning capability of DocShield is driven by our novel \textbf{Cross-Cues-aware Chain of Thought (CCT)} mechanism. Unlike standard MLLM prompting, CCT enforces a rigorous, six-stage iterative reasoning pipeline: (1) Knowledge Preparation, (2) Visual Cues Extraction, (3) Logical Cues Extraction, (4) Cross-Cues Validation \& Filtering, (5) Spatial Grounding, and (6) Report Synthesis. 
This design explicitly requires the model to jointly reason over visual artifacts and semantic inconsistencies, performing rigorous cross-validation before reaching a conclusion. 
Crucially, to eliminate reasoning hallucinations and ensure forensic faithfulness, we propose a specialized \textbf{Weighted Multi-Task Reward} function. Optimized via SFT and GRPO~\cite{shao2024deepseekmath}, 
it supervises the entire inference chain by jointly penalizing deviations in format compliance, spatial grounding, and explanation fidelity, enforcing a strict causal link between visual-logical anomalies and the final authenticity prediction.

However, existing document forgery datasets primarily provide coarse labels or limited annotations, which are insufficient to support multi-step, evidence-grounded reasoning. 
To bridge this gap, we construct \textbf{RealText-V1}, a multilingual, fine-grained benchmark specifically designed for text-centric forgery analysis with structured reasoning supervision. The dataset is generated via our \textbf{PR²} pipeline~\ref{fig:data_pipeline}, a hierarchical agentic framework comprising a \textbf{Perceiver}, \textbf{Reasoner}, and \textbf{Reviewer}, ensuring high-quality, consistent annotations across detection, grounding, and explanation tasks.
Our main contributions are summarized as follows:
\begin{itemize}
    \item We propose DocShield, a unified generative framework that reformulates forensic detection, spatial grounding, and explanation into a joint evidence-grounded co-reasoning task. Its reasoning capability is unlocked by the novel Cross-Cues-aware Chain-of-Thought (CCT) mechanism, enabling robust visual-logical cross-validation.
    
    \item To mitigate error propagation across decoupled stages and eliminate hallucinations, we design a Weighted Multi-Task Reward mechanism. Leveraging GRPO, it enforces strict joint alignment across format, grounding, and explanation objectives, ensuring predictions rely on consistent evidence rather than cascading errors.

    \item We construct RealText-V1, the first benchmark featuring fine-grained, multi-task annotations for text-centric analysis, generated via a novel multi-agent pipeline. Extensive experiments demonstrate that DocShield achieves state-of-the-art performance and superior zero-shot robustness across multiple benchmarks.
\end{itemize}
\section{Related Works}
\label{sec:formatting}

\subsection{Forgery Analysis in Text-Centric Images}
\label{subsec:forgery_detection}

Conventionally, the analysis of manipulated text in images has been framed as Tampered Text Detection~\cite{ahmed2014forgery,wang2022tampered,song2024cross,bertrand2015conditional}, focusing primarily on binary classification. The recent shift toward Explainable Tampered Text Detection~\cite{qu2024textsleuth} introduces interpretability, yet often compromises detection robustness and fine-grained grounding. These early methods provide foundational forensic insights for reasoning-based approaches. For example, DTD~\cite{qu2023towards} and CAFTB-Net~\cite{song2024cross} effectively leverage visual and frequency artifacts (e.g., boundary inconsistencies), while TextSleuth~\cite{qu2024textsleuth} explores the potential of MLLMs for generating visual-based explanations. 
However, these methods are predominantly driven by visual artifacts, often overlooking the semantic inconsistencies and logical flaws hidden within the text.

\subsection{Explainable Forgery Analysis with MLLMs}
\label{subsec:explainable_analysis}

The emergence of MLLMs has advanced explainable forensics along two paradigms. The first, prompt-driven analysis, leverages pre-trained MLLMs via zero-shot inference (e.g., GPT-4o) or prompt tuning (e.g., AntifakePrompt~\cite{chang2024antifakeprompt}) but is limited to image-level classification and lacks fine-grained grounding. The second stream employs segmentation-based architectures like FakeShield~\cite{xu2025fakeshield} and LEGION~\cite{kang2026legion}, yet relies on decoupled, multi-stage pipelines that first generate coarse textual descriptions and then produce segmentation masks in separate modules. This cascaded design suffers from error propagation and impedes cross-modal reasoning. Although LEGION~\cite{kang2026legion} explored physical law violations in images, it is biased toward fully synthetic natural images. Most approaches~\cite{lai2024lisa,rasheed2024glamm,huang2024sida} prioritize visual cues while neglecting logical and semantic contradictions in the text, highlighting the need for a unified, end-to-end framework capable of deep cross-modal forgery analysis.
\section{Methodology}
\label{sec:methodology}

Inspired by recent advancements in MLLM-based forgery analysis~\cite{lai2024lisa, xu2025fakeshield,kang2026legion,nguyen2024editscout,li2024forgerygpt,sun2024forgerysleuth,gao2025fakereasoning}, we introduce DocShield, a unified reasoning framework designed to address the limitations discussed in Section~\ref{sec:intro}. The foundation of DocShield is its end-to-end architecture, which integrates detection, grounding, and explanation into a single, cohesive process. This unified design directly addresses the fragmentation in prior work, allowing for synergistic reasoning that improves overall performance.
Building upon this architecture, we introduce a novel Cross-Cues-aware Chain of Thought (CCT) mechanism. 
This core component moves beyond analyzing clues in isolation, empowering DocShield to fuse evidence from both visual artifacts and logical reasoning to form a more holistic and reliable judgment.
To ensure these complementary capabilities are optimized jointly, we employ a weighted multi-task reward function trained with the GRPO algorithm, promoting both stable and high-performing outputs.

\subsection{Unified Generative Formulation}
\label{subsec:overview}

In this section, we define~\textbf{text-centric forgery analysis}, a unified challenge that requires models to simultaneously detect, ground, and explain manipulations in any image containing text.
To achieve this, we reformulate the entire challenge into a single, joint generative task.
Specifically, the model learns to generate a single textual analysis report, denoted as \(R_{\text{analysis}}\), that encompasses all three outputs: a textual classification (e.g., "HIGH INDICATION OF FORGERY")~for detection (\(S_{\text{det}}\)), a coordinate string (e.g., "[x1, y1, ...x4, y4]") for grounding (\(L_{\text{coords}}\)), and a detailed rationale for explanation (\(E_{\text{rationale}}\)). 
This unified process, where the entire report is generated by a single MLLM decoder \(\mathcal{D}\) conditioned on the input image \(I\) and task prompt \(T\) processed by an encoder \(\mathcal{E}\), can be expressed as:
\begin{equation}
    R_{\text{analysis}} \leftarrow \mathcal{D}(\mathcal{E}(I, T))
    \label{eq:unified_generation}
\end{equation}
The structured output also enables numerous downstream applications, such as automatically generating professional, multi-modal forensic documents in Fig.~\ref{fig:framework_overview}. 

\subsection{Cross-Cues-aware Chain of Thought (CCT)}
\label{subsec:cct}

To systematically validate visual and logical anomalies, we introduce the CCT mechanism, which correlates multi-modal evidence to establish rigorous, evidence-based forensic reasoning, thereby mitigating the risk of hallucinated explanations. As illustrated in Figure~\ref{fig:framework_overview} and detailed in Algorithm~\cref{alg:DocShield} (Appendix~\textbf{C} for additional visualizations), this cognitive workflow operates through a structured six-stage pipeline:

\noindent\textbf{Knowledge Preparation.(Implicit)}
This initial stage establishes a holistic semantic and forensic understanding of the input image. First, the model performs implicit image captioning to capture the global context (e.g., identifying a ``vintage baseball player card''). This semantic prior triggers the retrieval of domain-specific world knowledge, such as domain rules (e.g., baseball batting average calculations) and historical context. Concurrently, internal forensic priors are activated, mapping the scene context to probable manipulation typologies (e.g., printed anomalies, copy-move splicing). Then, an Optical Character Recognition (OCR) task extracts the precise spatial coordinates and text content.

\noindent\textbf{Visual Cues Extraction.}
This stage isolates visual anomalies through an implicit, dual-scale reasoning process, requiring no external cropping operators. The model first evaluates global features (e.g., printing quality, lighting coherence) before transitioning to a pixel-level inspection of localized high-risk regions. As illustrated in Figure~\ref{fig:framework_overview}~(2), this process identifies font rendering inconsistencies in the player name ``AL MATTIN'', detecting edge aliasing and unnatural chromatic aberrations (marked by the \textcolor{red}{red} cross). However, relying solely on visual cues inherently leaves blind spots. Sophisticated manipulations—such as altered statistic tables, logos, or QR codes—can maintain high visual fidelity, deceiving the model into initially accepting them as pristine (\textcolor{green}{green} checkmarks). This underscores the fundamental limitation of unimodal visual analysis.

\noindent\textbf{Logical Cues Extraction.}
Moving beyond low-level visual analysis, this stage performs a higher-level semantic check. By leveraging both the semantic world knowledge and forensic priors activated in Stage 1, the model identifies deeper contradictions within the broader text-centric context. As illustrated in Figure~\ref{fig:framework_overview}~(3), the model is guided to detect anomalies along two dimensions. \textit{Internal Consistency Verification} evaluates mathematical calculations and sequential impossibilities. For instance, the model flags strong math errors where the stated batting average (\textit{.381}) defies the actual calculation (135/480 $\approx$ \textit{.281}), and exposes a profound timeline contradiction where a player starting a pro career in '92 but born in '87 would be \textit{only 5 years old}. In parallel, \textit{Scenario Logic Verification} assesses external plausibility against historical common sense, identifying medium-confidence layout anachronisms such as a modern \textit{QR} code appearing on a trading card with a \textit{2001} copyright. These deep-seated logical flaws are distinctively highlighted in the top-level visualization of this stage.

\noindent\textbf{Cross-Cues Validation \& Filtering.}
This stage cross-validates the extracted visual and logical cues for rigorous evidence synthesis. To ensure robustness, DocShield utilizes a \textit{complementary validation} mechanism (Figure~\ref{fig:framework_overview}~(4)), where logical reasoning (Stage 3) compensates for the visual blind spots (Stage 2). For instance, while manipulated statistic tables or logos may appear visually intact, logical analysis exposes their underlying semantic contradictions. By filtering isolated, low-confidence findings, this convergence consolidates all validated anomalies—ranging from visual splicing (\textcolor{MyPurple}{'TT'}) to complex math errors (H/AB) and historical anachronisms (QR code). Ultimately, this constructs a comprehensive evidence chain that reliably localizes all tampered regions.

\noindent\textbf{Grounding.}
This stage precisely localizes the tampered regions by reactivating the fine-grained OCR output from Stage 1. Through targeted semantic retrieval, the model matches the cross-validated anomalies against the structured OCR data. This semantic-to-spatial mapping directly links the abstract reasoning back to the image coordinates, outputting the exact bounding boxes (e.g., \texttt{[x0, y0, x1, y1]}) of the manipulated text.

\noindent\textbf{Report Synthesis.}
Finally, the model internally consolidates all validated findings: the authenticity prediction, spatial coordinates, and visual-logical rationale. Triggered by a specialized \texttt{<report>} token, it synthesizes this comprehensive evidence into a structured JSON format, ready for downstream forensic applications.
\begin{algorithm}[t]
\caption{Cross-Cues-aware Chain of Thought (CCT) Process}
\label{alg:DocShield}
\begin{algorithmic}[1]
\Statex \textbf{Input:} Image $I$
\Statex \textbf{Output:} Structured Analysis Report $R$
\Statex 

\Statex \textbf{Stage 1: }
\State $C_{\text{caption}} \gets \text{ImplicitImageCaptioning}(I)$
\State $K_{\text{world}} \gets \text{ActivateWorldKnowledge}(C_{\text{caption}})$
\State $K_{\text{forensic}} \gets \text{ActivateForensicKnowledge}(C_{\text{caption}})$
\State $D_{\text{ocr}} \gets \text{ImplicitExtractTextInfo}(I)$

\Statex \textbf{Stage 2: }
\State $Cues_{\text{global}} \gets \text{AnalyzeGlobalConsistency}(I, K_{\text{forensic}})$
\State $Cues_{\text{local}} \gets \text{InspectLocalAnomalies}(I, K_{\text{forensic}})$
\State $V_{\text{cues}} \gets \text{AggregateCues}(Cues_{\text{global}}, Cues_{\text{local}})$

\Statex \textbf{Stage 3:}
\State $Cues_{\text{internal}} \gets \text{VerifyInternalConsistency}(D_{\text{ocr}})$
\State $Cues_{\text{scenario}} \gets \text{VerifyScenarioLogic}(D_{\text{ocr}}, K_{\text{world}})$
\State $L_{\text{cues}} \gets \text{AggregateCues}(Cues_{\text{internal}}, Cues_{\text{scenario}})$

\Statex \textbf{Stage 4: }
\State $All_{\text{cues}} \gets V_{\text{cues}} \cup L_{\text{cues}}$
\State $Scores \gets \text{InternalScoringMechanism}(All_{\text{cues}})$
\State $HighValue_{\text{cues}} \gets \text{FilterByScore}(All_{\text{cues}}, Scores)$

\Statex \textbf{Stage 5:}
\State $TamperedRegions \gets \emptyset$
\ForAll{$cue \in HighValue_{\text{cues}}$}
    \State $matched\_text \gets \text{SemanticRetrieval}(cue, D_{\text{ocr}})$
    \State $TamperedRegions \gets TamperedRegions \cup \text{GetCoordinates}(matched\_text)$
\EndFor

\Statex \textbf{Stage 6:}
\State $Verdict \gets \text{DetermineFinalVerdict}(HighValue_{\text{cues}})$
\State $Rationale \gets \text{ConsolidateEvidence}(HighValue_{\text{cues}})$
\State $R \gets \text{GenerateReport}(Verdict, Rationale,$
\Statex \hspace{\algorithmicindent} $TamperedRegions)$

\State \textbf{return} $R$
\end{algorithmic}
\end{algorithm}

\subsection{Weighted Multi-Task Reward Function}
\label{subsec:reward_function}

Our composite reward function $R_{\text{total}}$ is a weighted sum of three distinct components, inspired by~\cite{huang2025lavcot}. Each component targets a critical aspect of the final multimodal reasoning output:
\begin{equation}
    R_{\text{total}} = \lambda_f R_{\text{format}} + \lambda_g R_{\text{grounding}} + \lambda_e R_{\text{explanation}}.
\end{equation}
Let $y$ be the model's generated report and $\hat{y}$ be the ground-truth report. In our experiments, these coefficients are empirically set to $\lambda_f = 0.15$, $\lambda_g = 0.75$, and $\lambda_e = 0.1$. Each component is defined as follows:

\begin{itemize}
    \item \textbf{Format Adherence Reward ($R_{\text{format}}$):} To encourage well-structured outputs, this reward evaluates the presence of key structural tags. 
    
    Let $\mathcal{T} = \{\text{\texttt{<think>}}, \text{\texttt{</think>}}, \text{\texttt{<report>}}, \text{\texttt{</report>}}\}$ be the set of required structural tokens. The reward is the proportion of tags correctly generated in output $y$:
    \begin{equation}
        R_{\text{format}} = \frac{1}{|\mathcal{T}|} \sum_{t \in \mathcal{T}} \mathbb{I}(t \in y),
    \end{equation}
    where $\mathbb{I}(\cdot)$ is the indicator function.

    \item \textbf{Grounding Accuracy Reward ($R_{\text{grounding}}$):} This is a multi-faceted reward explicitly evaluating detection and spatial localization. Let $c$ and $\hat{c}$ denote the predicted and ground-truth binary authenticity labels, respectively. Let $B$ and $\hat{B}$ represent the set of predicted and ground-truth spatial bounding boxes. The reward aggregates three sub-metrics:
    \begin{equation}
        R_{\text{grounding}} = w_c R_{\text{cls}} + w_n R_{\text{num}} + w_{\text{iou}} R_{\text{iou}}.
    \end{equation}
    
    \begin{enumerate}
        \item \textit{Detection Accuracy ($R_{\text{cls}}$):} A binary reward for correct image-level classification:
        \begin{equation}
            R_{\text{cls}} = \mathbb{I}(c = \hat{c}).
        \end{equation}
        
        \item \textit{Object Count Precision ($R_{\text{num}}$):} This incentivizes the model to predict the exact number of tampered regions, penalizing both hallucinations and omissions:
        \begin{equation}
            R_{\text{num}} = 
            \begin{cases} 
                0.5 \cdot \mathbb{I}(|B| = |\hat{B}|), & \text{if } \hat{c} = \text{forged} \\
                0.5 \cdot \mathbb{I}(|B| = 0),     & \text{if } \hat{c} = \text{authentic}
            \end{cases}
        \end{equation}
        
        \item \textit{Localization IoU Score ($R_{\text{iou}}$):} This assesses spatial grounding accuracy. For forged images ($|\hat{B}| > 0$), we compute the mean Intersection over Union (mIoU). For each ground-truth box $\hat{b} \in \hat{B}$, we find its maximum IoU against all predicted boxes $b \in B$:
        \begin{equation}
            \text{mIoU} = \frac{1}{|\hat{B}|} \sum_{\hat{b} \in \hat{B}} \max_{b \in B} \text{IoU}(\hat{b}, b).
            \label{eq:miou_calc}
        \end{equation}
        To penalize poor localization and reward high precision, this mIoU score is mapped to a discrete reward value $R_{\text{iou}}$ (for authentic images, $R_{\text{iou}}$ naturally evaluates to $0.0$):
        \begin{equation}
            R_{\text{iou}} =
            \begin{cases}
                0.6, & \text{if mIoU} > 0.8 \\
                0.4, & \text{if } 0.5 \le \text{mIoU} \le 0.8 \\
                0.0, & \text{otherwise}
            \end{cases}
            \label{eq:riou_tiered}
        \end{equation}
    \end{enumerate}
    
    \item \textbf{Explanation Quality Reward ($R_{\text{explanation}}$):} To evaluate the semantic fidelity of the generated logical reasoning, we compute the cosine similarity between the sentence embeddings of the generated explanation $y$ and the ground-truth rationale $\hat{y}$. Utilizing a pre-trained sentence embedding model $\phi(\cdot)$ (e.g., Qwen3-Embedding-4B), the semantic alignment is measured as:
    \begin{equation}
        R_{\text{explanation}} = \text{sim}(\phi(y), \phi(\hat{y})).
    \end{equation}
\end{itemize}

\section{Datasets}
\label{sec:dataset}

\subsection{Motivation}
\label{subsec:motivation_principles}
\begin{figure*}[t]
  \centering
  \includegraphics[width=\textwidth, height=0.38\textwidth]{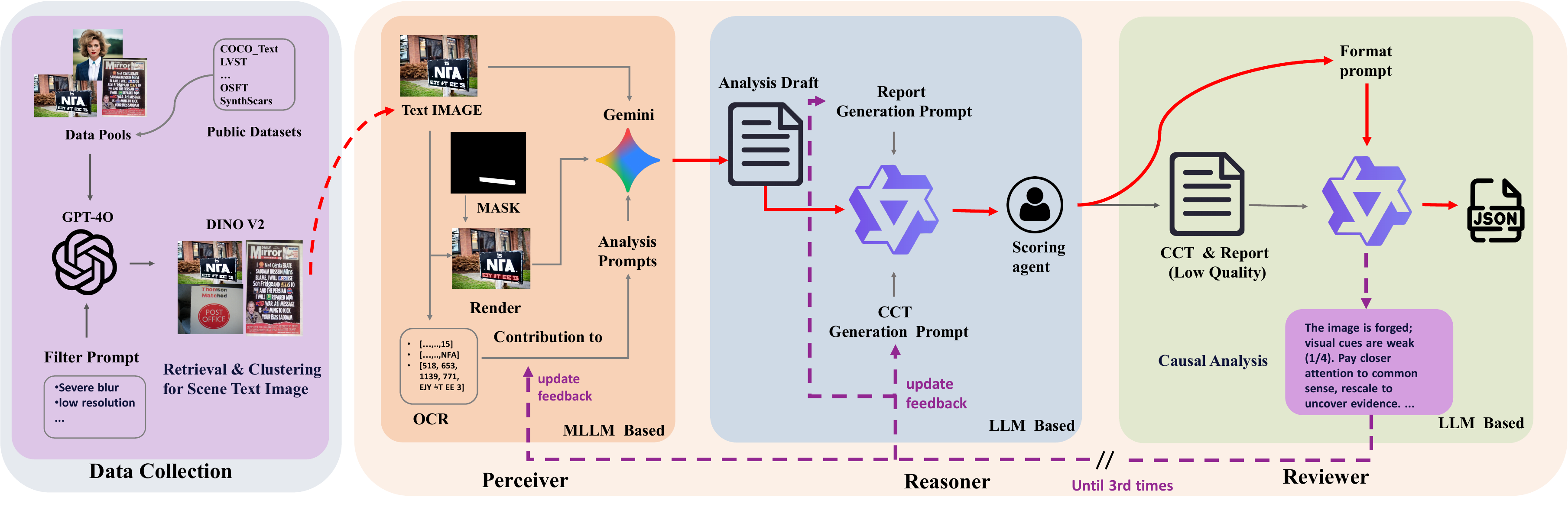}
  \caption{The architecture of our \textbf{PR²} (\textbf{P}erceiver, \textbf{R}easoner, \textbf{R}eviewer) pipeline. After an initial data collection stage, our multi-agent system generates annotations through a collaborative, iterative process. The Perceiver drafts an analysis, the Reasoner structures it to target CCT \& analysis report, and the Reviewer validates its quality, initiating a \textbf{refinement loop} if necessary. This cycle, indicated by the solid~\textcolor{red}{$\bm{\rightarrow}$} and dashed~\textcolor{MyPurple}{$\bm{\dashleftarrow}$} feedback loops, ensures the final output is a high-fidelity, structured JSON annotation.}
  \Description{Architecture diagram of the PR² (Perceiver, Reasoner, Reviewer) data curation pipeline. The figure shows the multi-agent system workflow: data collection stage, followed by collaborative annotation generation where the Perceiver drafts analysis, the Reasoner structures it for CCT and report, and the Reviewer validates quality with iterative refinement loops indicated by solid red and dashed purple arrows, producing structured JSON annotations.}
  \label{fig:data_pipeline}
\end{figure*}

A comprehensive benchmark for explainable text-centric forgery analysis is notably absent. While existing datasets have made valuable contributions, each addresses only a fragment of the problem. T-IC13~\cite{wang2022detecting} and T-SROIE~\cite{wang2022tampered} provide foundational forged samples but are limited by scale and class imbalance. 
OSFT~\cite{qu2024revisiting} enhances data diversity with diffusion models; however, none of these benchmarks provide the fine-grained, explanatory annotations required for deep visual-logical reasoning. Even the most recent proposal, ETTD~\cite{qu2024textsleuth}, maintains a visual-only focus and remains publicly unavailable.

\begin{table}[t]
  \centering
  \caption{RealText-V1 is the first benchmark with comprehensive annotations for text-centric forgery analysis: textline-level annotations (T-line), multilingual support (M-Lang), detection (Det), grounding (Mask), and explanation (Expl).}
  \label{tab:dataset_comparison_final_complete}
  \setlength{\tabcolsep}{2.5pt} 
  \begin{tabular}{lcccccc}
    \toprule
    \textbf{Dataset} & \textbf{Total} & \textbf{T-line} & \textbf{M-Lang} & \textbf{\makecell{Det}}  & \textbf{Mask}  & \textbf{Expl}\\
    \midrule
    T-IC13~\cite{wang2022detecting}     & 462       & \checkmark & \ding{55} & \ding{55} & \ding{55} & \ding{55} \\
    T-SROIE~\cite{wang2022tampered}     & 986       & \checkmark & \ding{55} & \ding{55} & \ding{55} & \ding{55} \\
    OSFT~\cite{qu2024revisiting}        & 2,938     & \checkmark & \ding{55} & \ding{55} & \ding{55} & \ding{55} \\
    DocTamper~\cite{qu2023towards}        & 170,000    & \ding{55}  & \ding{55} & \ding{55} & \checkmark & \ding{55} \\
    \midrule
    \textbf{RealText-V1 (Ours)}         & \textbf{5,397} & \checkmark & \checkmark & \checkmark & \checkmark & \checkmark \\
    \bottomrule
  \end{tabular}
\end{table}
\subsection{Data Curation Pipeline}
\label{subsec:data_curation}

To address these limitations, we developed \textbf{PR²}, a hierarchical, feedback-driven multi-agent pipeline (Fig.~\ref{fig:data_pipeline}) to construct our dataset, \textbf{RealText-V1}. 
As a foundation, we aggregate diverse samples from public benchmarks~\cite{veit2016cocotext,zhang2019icdarrects,sun2019icdar,kang2026legion,qu2024revisiting}, applying rigorous visual-semantic filtering via DINOv2~\cite{oquab2023dinov2} and GPT-4o~\cite{openai2024gpt4o}. For each retained image, PR² is initialized with its fused manipulation mask and OCR transcript. While the architecture supports flexible model swapping, our current configuration is empirically optimized to strictly balance annotation quality and computational cost. The data subsequently flows through three specialized agents:

\noindent\textbf{The Perceiver (Holistic Forgery Analysis):} 
Guided by the designed analysis prompt (detailed in Appendix \textbf{B}), the Perceiver performs a comprehensive assessment of the highlighted regions on the fused mask-image. This role is powered by Gemini-2.5-Pro~\cite{google2024gemini2.5}, chosen specifically because it empirically excelled in fine-grained visual reasoning tasks during our benchmarking. It generates a preliminary report detailing both visual artifacts and logical anomaly cues. While less structured than the final CCT, this initial draft provides the foundational raw material for deep reasoning.

\noindent\textbf{The Reasoner (Structured Refinement \& Scoring):} 
The Reasoner (powered by Qwen3-Max) ingests the Perceiver's raw text. Guided by specialized prompts, it executes two sequential tasks: first, it structures the raw analysis into a formalized CCT format and a coherent forensic report; second, it assigns a preliminary quality score based on global dimensions, including format integrity, conclusion consistency, and logical completeness.

\noindent\textbf{The Reviewer (Iterative Verification \& Refinement):} The Reviewer orchestrates the quality control loop by evaluating the structured annotations against a strict threshold (e.g., $\ge$ 98/100). For sub-optimal outputs, it conducts a root-cause analysis to generate feedback, initiating an iterative refinement cycle with the prior modules (capped at three iterations). Validated annotations are ultimately serialized into JSON to facilitate the multi-task reward calculation (Section~\ref{subsec:reward_function}).

\subsection{Dataset Statistics and Comparison}
\label{subsec:dataset_stats}
As detailed in Table~\ref{tab:dataset_comparison_final_complete}, we present RealText-V1, a large-scale benchmark comprising 5,397 images, specifically designed to advance research in explainable text-centric forgery analysis. Its primary distinction lies in the unparalleled comprehensiveness of its fine-grained annotations, successfully unifying detection, spatial grounding, and visual-logical explanation into a single framework. 

Crucially, RealText-V1 is highly multilingual, encompassing texts across diverse languages. This extensive linguistic coverage ensures that models trained on this benchmark can generalize across different character sets, typography styles, and cultural contexts. Furthermore, the dataset features a challenging subset of dense-text images, deliberately included to facilitate the rigorous evaluation of model robustness under complex real-world conditions. As such, RealText-V1 represents a vital and versatile new resource for the multimedia and AI document security community.

\section{Experiments}
\label{sec:experiments}
\subsection{Experimental Setup}
\label{subsec:setup}

\noindent\textbf{Implementation Details.}  
DocShield is initialized with Qwen2.5-VL-7B~\cite{Qwen2.5-VL}. Training proceeds in two stages: (1) Supervised Fine-Tuning (SFT) for 30 epochs using LoRA~\cite{hu2022lora} ($r=32, \alpha=64$) applied to all attention projection layers, with a learning rate of $1 \times 10^{-4}$; and (2) a GRPO-based reinforcement learning alignment stage for 10 epochs with a reduced learning rate of $1 \times 10^{-5}$. Input images are resized to $1344 \times 896$, and the vocabulary is extended with special tokens \texttt{<think>} and \texttt{<report>}. Training is conducted on 8 NVIDIA A100 GPUs.  
For practical deployability, inference is evaluated on a single RTX 4090 (24GB) GPU, operating in BF16 precision with a peak VRAM usage of $\approx 16$GB. Processing a high-resolution image requires roughly 35 seconds, reflecting a deliberate design trade-off: we prioritize high-resolution input and multi-stage CCT reasoning depth over real-time speed to ensure precise detection of subtle artifacts and robust forensic reasoning.  
All primary training and evaluation are performed on the RealText-V1 benchmark. Out-of-domain generalization to sparse scene text is assessed in a zero-shot setting on T-IC13, while robustness under dense-text scenarios is evaluated on the challenging T-SROIE benchmark.

\begin{table*}[t]
  \centering
  \caption{Comprehensive performance comparison across three benchmarks: proposed RealText, T-IC13 (\textbf{zero-shot}) and T-SROIE (\textbf{dense text for robustness}), DocShield demonstrates superior performance, strong generalization and robustness. All metrics are in percentage (\%). Acc: Accuracy, mIOU: mean Intersection over Union, CSS: Cosine Similarity Score, BS(F1): BERTScore F1. M-F1: the macro-average of its Detection F1, Grounding mF1, and Explanation BS(F1). Best results are in \textbf{bold}, second best are \underline{underlined}, '-': Model unreleased and result not reported.}
  \label{tab:main_results}
  \resizebox{\textwidth}{!}{%
    \begin{tabular}{@{}l|cc|cc|cc|c|cc|cc|cc|c|cc|cc|cc|c@{}}
      \toprule
      \multirow{3}{*}{\textbf{Method}} & \multicolumn{7}{c|}{\textbf{RealText}} & \multicolumn{7}{c|}{\textbf{T-IC13}} & \multicolumn{7}{c}{\textbf{T-SROIE}} \\
      \cmidrule(l){2-8} \cmidrule(l){9-15} \cmidrule(l){16-22}
      
       & \multicolumn{2}{c|}{Detection} & \multicolumn{2}{c|}{Grounding} & \multicolumn{2}{c|}{Explanation} & \multirow{2}{*}{\shortstack{M-F1}}
       & \multicolumn{2}{c|}{Detection} & \multicolumn{2}{c|}{Grounding} & \multicolumn{2}{c|}{Explanation} & \multirow{2}{*}{\shortstack{M-F1}}
       & \multicolumn{2}{c|}{Detection} & \multicolumn{2}{c|}{Grounding} & \multicolumn{2}{c|}{Explanation} & \multirow{2}{*}{\shortstack{M-F1}} \\
      \cmidrule(l){2-3} \cmidrule(l){4-5} \cmidrule(l){6-7} 
      \cmidrule(l){9-10} \cmidrule(l){11-12} \cmidrule(l){13-14} 
      \cmidrule(l){16-17} \cmidrule(l){18-19} \cmidrule(l){20-21} 
      
       & Acc & F1 & mIOU & mF1 & CSS & BS(F1) & 
       & Acc & F1 & mIOU & mF1 & CSS & BS(F1) & 
       & Acc & F1 & mIOU & mF1 & CSS & BS(F1) & \\ 
       \midrule
       
      \multicolumn{21}{l}{\textit{\textbf{Group 1: State-of-the-Art MLLM-based Frameworks}}} \\
      FakeShield~\cite{xu2025fakeshield} & 58.2 & 55.6 & \underline{30.5} & 12.4 & 66.6 & 53.8 & 40.6 & 36.9 & 32.3 & \underline{23.8} & \underline{26.2} & 57.5 & 52.7 & 37.1 & 14.7 & 25.7 & 0.5 & \underline{1.9} & 51.7 & 52.9 & 26.8  \\
      TextSleuth-7B~\cite{qu2024textsleuth} & - & - & - & - & - & - & - & \underline{88.4} & - & - & - & - & - & - & - & - & - & - & - & - & -  \\
      \midrule
      \multicolumn{21}{l}{\textit{\textbf{Group 2: Powerful General-Purpose MLLMs}}} \\
      gpt-4o-o3-0416~\cite{openai2024gpt4o} & \underline{81.5} & \underline{84.5} & 8.1 & 15.1 & \underline{75.4} & 67.7 & \underline{55.8} & 85.7 & \underline{90.1} & 22.9 & 13.5 & 69.1& 65.7 & \underline{56.4} & 84.5 & 91.6 & \underline{1.3} & 1.3 & 73.3 & 65.5 & 52.8  \\
      Gemini-2.5-Pro~\cite{google2024gemini2.5} & 78.7 & 83.2 & 5.3 & 10.8 & 74.8 & \underline{67.8}  & 53.9 & 80.4 & 87.0 & 17.9 & 12.1 & \underline{76.9} & \underline{68.7}  & 55.9 & \textbf{99.9} & \textbf{99.9} & 0.2 & 1.2 & \underline{79.8} & \underline{69.8} & \underline{57.0}  \\
      InternVL-3.5-8B~\cite{wang2025internvl3.5} & 59.9 & 70.0 & 7.4 & 9.4 & 75.3 & 67.1 & 48.8 & 63.1 & 73.5 & 15.9 & 9.8 & 71.4 & 66.4 & 49.9 & 86.4 & 92.7 & 0.1 & 1.0 & 74.7 & 67.5 & 53.7  \\
      qwen3-vl-8B~\cite{qwen3technicalreport} & 63.4 & 72.2 & 7.3 & 10.7 & 72.8 & 66.2 & 49.7 & 80.3 & 88.0 & 12.7 & 14.5 & 72.1 & 66.5  & 56.3 & \textbf{99.9} & \textbf{99.9} & 0.0 & 1.0 & 75.9 & 66.0 & 55.6  \\
      DeepSeekVL-7B~\cite{deepseek2024deepseekvl} & 39.8 & 18.2 & 5.1 & 6.7 & 38.1 & 21.3 & 15.4 & 39.1 & 35.5 & 5.0 & 11.3 & 37.2 & 28.9 & 25.2 & 11.3 & 20.3 & 0.0 & 0.5 & 60.6 & 53.7 & 24.8 \\
      Qwen2.5-VL-7B~\cite{Qwen2.5-VL} & 43.2 & 27.8 & 17.8 & \underline{15.5} & 66.0 & 65.8 & 36.3 & 38.2 & 38.5 & 15.1 & 9.0 & 64.4 & 65.5 & 37.7 & 6.1 & 11.5 & 0.0 & 0.7 & 65.4 & 66.3  & 26.2  \\
      \midrule
      \rowcolor{gray!15}
      \textbf{DocShield (Ours)} & \textbf{91.4} & \textbf{93.2} & \textbf{32.4} & \textbf{36.7} & \textbf{85.0}&\textbf{76.9} & \textbf{68.9} & \textbf{91.2} & \textbf{93.2} & \textbf{44.5} & \textbf{67.6} & \textbf{86.7}&\textbf{78.5} & \textbf{79.8} & \underline{98.8} & \underline{99.4} & \textbf{9.1} & \textbf{11.0} & \textbf{84.9}&\textbf{77.0} & \textbf{62.5}  \\
      \bottomrule
    \end{tabular}%
  }
\end{table*}
\noindent\textbf{Evaluation Metrics.}
We evaluate the pipeline across three task components: (1) Detection is measured by standard Accuracy and F1-Score; (2) Spatial Grounding is assessed via pixel-level mean Intersection over Union (mIoU) and mean F1-Score (mF1), following the TruFor protocol~\cite{guillaro2023trufor}; and (3) Explanation quality is quantified using Cosine Similarity Score (CSS) and BERTScore~\cite{zhang2020bertscore}. Finally, we report the Macro-Average F1-Score (M-F1) across all three tasks as a unified performance indicator.
\begin{table}[h!]
  \centering
  \caption{Ablation study on the RealText-V1 Test set. Both the CCT mechanism and the weighted multi-task reward are shown to be critical for the performance of DocShield. All metrics are in percentage (\%). Best results are in \textbf{bold}.}
  \label{tab:ablation}
  \resizebox{\columnwidth}{!}{%
    \begin{tabular}{@{}l|cc|cc|cc|c@{}}
      \toprule
      \multirow{2}{*}{\textbf{Configuration}} & \multicolumn{2}{c|}{\textbf{Detection}} & \multicolumn{2}{c|}{\textbf{Grounding}} & \multicolumn{2}{c|}{\textbf{Explanation}} & \multirow{2}{*}{\shortstack{\textbf{M-F1}}} \\
      \cmidrule(l){2-3} \cmidrule(l){4-5} \cmidrule(l){6-7}
       & Acc & F1 & mIOU & mF1 & CSS & BS(F1) & \\
      \midrule
      \textbf{DocShield (Full)} & \textbf{91.4} & \textbf{93.2} & \textbf{32.4} & \textbf{36.7} & \textbf{85.0} & \textbf{76.9} & \textbf{68.9} \\
      \midrule
      \textit{w/o} GRPO (SFT only) & 89.4 & 90.3 & 27.6 & 31.1 & 81.9 & 73.2 & 64.9 \\
      \textit{w/o} CCT & 51.7 & 32.6 & 21.5 & 18.2 & 71.3 & 68.1 & 39.6 \\
      \midrule
      \textit{w/o} format reward & 91.2 & 92.6 & 32.1 & 36.0 & 83.8 & 75.4 & 68.0 \\
      \textit{w/o} grounding reward & 90.0 & 91.2 & 28.6 & 32.2 & 84.7 & 76.5 & 66.8 \\
      \textit{w/o} explanation reward & 91.0 & 93.0 & 31.7 & 36.3 & 83.5 & 75.1 & 67.9 \\
      \bottomrule
    \end{tabular}%
  }
\end{table}

\noindent\textbf{Main Results.}
As shown in Table~\ref{tab:main_results}, DocShield consistently outperforms both specialized forensic models and general-purpose MLLMs, achieving the highest M-F1 scores across RealText-V1, T-IC13, and T-SROIE benchmarks. This establishes a new state-of-the-art in unified text-centric forgery analysis.  
The model demonstrates strong zero-shot generalization, achieving an M-F1 of 79.8\% on T-IC13 compared to 56.4\% for GPT-4o, and maintains robust performance on the challenging T-SROIE dataset. DocShield also delivers competitive results relative to traditional forensic methods (Appendix~\textbf{A}), and surpasses the specialist DTD model on T-IC13 grounding (mF1 67.6\% vs. 45.4\%).  
Importantly, our 7B-parameter DocShield model outperforms considerably larger proprietary systems and MLLM baselines, indicating that its superior performance arises from architectural and training innovations rather than model scale. Qualitative examples highlighting visual artifact localization and logical reasoning are shown in Fig.~\ref{fig:qualitative}, with additional comparisons provided in Appendix~\textbf{D}.

\subsection{Ablation Studies}
\begin{figure*}[t]
  \centering
  \includegraphics[width=\textwidth]{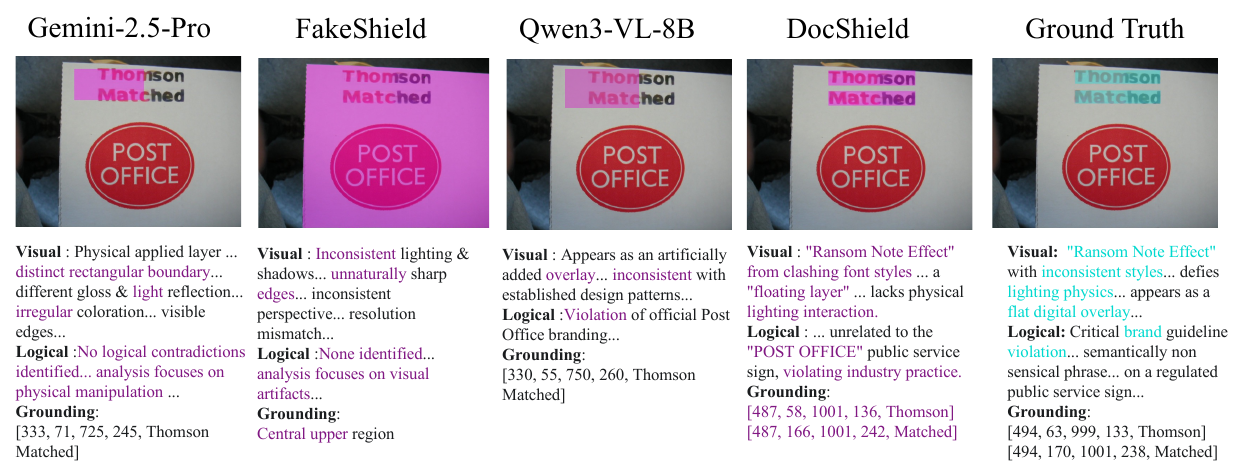}
  \caption{Qualitative comparison of artifact grounding and explanations across different methods. DocShield demonstrates superior performance, accurately identifying both visual artifacts and logical cues. Shaded regions indicate the localized tampered areas.}
  \Description{Qualitative comparison figure showing artifact grounding and explanation results across different methods on example forged document images. DocShield results demonstrate superior performance in identifying visual artifacts and logical cues, with shaded regions indicating localized tampered areas compared to other MLLM baselines.}
  \label{fig:qualitative}
\end{figure*}

\noindent\textbf{Component Analysis: CCT and Weighted Multi-Task Reward.}  
We first evaluate the contribution of the CCT mechanism and the GRPO alignment stage (Table~\ref{tab:ablation}). Removing CCT causes a substantial drop in performance, reducing M-F1 by 29.3\% and Detection F1 by over 60\%, highlighting its essential role in evidence-based reasoning. Similarly, training with SFT alone without GRPO alignment degrades M-F1 by 4.0\%. Analysis of the weighted multi-task reward further shows that the grounding reward is critical for discriminative performance, with its removal lowering M-F1 by 2.1\% and impairing localization. In contrast, format and explanation rewards primarily influence the structural compliance and semantic quality of machine-readable rationales.

\noindent\textbf{Training Dynamics: Optimization for Efficiency.}  
Beyond absolute performance metrics, the GRPO alignment stage optimizes the model's inference policy. As shown in Appendix~\textbf{A} (Training Curves), the format reward converges rapidly, accompanied by a reduction in output entropy and mean generation length (from $\sim$4,100 to 3,400 tokens). This demonstrates that the weighted multi-task reward effectively penalizes verbose or uncertain outputs, guiding DocShield toward more decisive, computationally efficient reasoning while maintaining accuracy.

\noindent\textbf{Evaluating Reasoning Paradigm and Visual–Logical Fusion.}
To assess whether the model executes genuine reasoning rather than relying on linguistic priors, we introduce the FCLE baseline (\textbf{F}irst \textbf{C}lassify, \textbf{L}ast \textbf{E}xplain), implemented with Qwen2.5-VL-7B. Prompted to output a binary verdict before providing an explanation (Appendix~\textbf{B}), simulating the hallucination-prone behavior of conventional MLLMs. As shown in Table~\ref{tab:cct_ablation}, \textit{FCLE (Full)} achieves a Detection F1 of 56.1\%, substantially lower than DocShield (93.2\%). 

We further isolate the input modalities within the FCLE paradigm to expose the limitations of unimodal reasoning. Specifically,\textit{FCLE (Visual Only)} achieves a Detection F1 of 11.8\%, indicating insensitivity to subtle pixel-level anomalies. \textit{FCLE (Logical Only)} improves detection (F1: 46.5\%) but fails at spatial localization (Grounding mIOU: 5.1\%). These results indicate that while textual details provide important logical cues, they alone cannot guarantee accurate detection or precise spatial grounding, underscoring the necessity of structured, cross-modal reasoning.

\noindent\textbf{Impact of SFT and GRPO on Visual Reasoning.}  
As illustrated in Table~\ref{tab:cct_ablation}, comparing \textit{FCLE (Visual Only)} with the CCT variant lacking logical guidance (\textit{w/o Logical Cues}) demonstrates the effect of SFT and GRPO optimization. The \textit{w/o Logical Cues} variant achieves a Detection F1 of 81.2\%, compared to 11.8\% for \textit{FCLE (Visual Only)}, marking a significant absolute improvement. This empirically confirms that the training procedure equips the vision encoder to robustly capture physical tampering artifacts, rather than merely learning a new prompting format.

\noindent\textbf{Effect of CCT Components on Reasoning Fidelity.}  
As shown in Table~\ref{tab:cct_ablation}, targeted ablations highlight the importance of CCT's internal stages. Removing visual cues (\textit{w/o Visual Cues}) maintains a Detection F1 of 88.7\% but drops Grounding mIOU to 25.3\%, demonstrating that visual information is essential for precise localization. Omitting the \textit{Cross-Cues Validation} stage yields a Detection F1 of 91.6\%, yet explainability metrics degrade significantly (CSS: 81.3\%, BS(F1): 72.5\%), reducing the macro F1 to 64.3\%. This decoupling of detection accuracy from explanation fidelity confirms that cross-modal validation is critical: without it, the model produces plausible but ungrounded rationales and bounding boxes. Therefore, the Cross-Cues stage is vital for faithful, evidence-grounded reasoning. Additional visualizations of the reasoning chain are provided in Appendix~\textbf{C}.

\begin{table}[h]
\vspace{-1.0em}
\centering
\setlength{\tabcolsep}{2.5pt} 
\caption{\small Robustness Comparison under Extreme Perturbations. \textbf{Det-F1}: Detection F1. \textbf{M-F1}: Macro-average F1. Values in parentheses denote the performance drop relative to the No Distortion baseline. The most resilient method (smallest relative drop) for each metric is highlighted in \textcolor{red}{red}.}
\label{tab:robustness_single}
\resizebox{\linewidth}{!}{
\begin{tabular}{l|cc|cc|cc}
\toprule
\multirow{2}{*}{\textbf{Distortion}} & \multicolumn{2}{c|}{FakeShield} & \multicolumn{2}{c|}{Gemini-2.5-Pro} & \multicolumn{2}{c}{\cellcolor{gray!15}\textbf{DocShield}} \\
 & Det-F1 & M-F1 & Det-F1 & M-F1 & \cellcolor{gray!15}\textbf{Det-F1} & \cellcolor{gray!15}\textbf{M-F1} \\
\midrule
\textbf{No Distortion} & 55.6 & 40.6 & 83.2 & 53.9 & \cellcolor{gray!15}\textbf{93.2} & \cellcolor{gray!15}\textbf{68.9} \\
\midrule
Blur ($K$=15) & 49.3 \tiny{($-6.3$)} & 35.5 \tiny{($-5.1$)} & 73.3 \tiny{($-9.9$)} & 47.6 \tiny{($-6.3$)} & \cellcolor{gray!15}88.3 \tiny{(\textcolor{red}{$-4.9$})} & \cellcolor{gray!15}64.2 \tiny{(\textcolor{red}{$-4.7$})} \\
JPEG ($Q$=20) & 47.4 \tiny{($-8.2$)} & 32.7 \tiny{($-7.9$)} & 69.6 \tiny{($-13.6$)} & 48.8 \tiny{(\textcolor{red}{$-5.1$})} & \cellcolor{gray!15}85.9 \tiny{(\textcolor{red}{$-7.3$})} & \cellcolor{gray!15}61.3 \tiny{($-7.6$)} \\
Noise ($\sigma$=0.3) & 46.2 \tiny{($-9.4$)} & 31.9 \tiny{($-8.7$)} & 61.2 \tiny{($-22.0$)} & 36.2 \tiny{($-17.7$)} & \cellcolor{gray!15}84.7 \tiny{(\textcolor{red}{$-8.5$})} & \cellcolor{gray!15}61.1 \tiny{(\textcolor{red}{$-7.8$})} \\
Rotate ($10^{\circ}$) & 43.0 \tiny{($-12.6$)} & 30.3 \tiny{($-10.3$)} & 73.1 \tiny{($-10.1$)} & 47.2 \tiny{($-6.7$)} & \cellcolor{gray!15}89.8 \tiny{(\textcolor{red}{$-3.4$})} & \cellcolor{gray!15}63.4 \tiny{(\textcolor{red}{$-5.5$})} \\
Flip (Horiz.) & 50.8 \tiny{(\textcolor{red}{$-4.8$})} & 33.5 \tiny{($-7.1$)} & 67.3 \tiny{($-15.9$)} & 41.5 \tiny{($-12.4$)} & \cellcolor{gray!15}87.3 \tiny{($-5.9$)} & \cellcolor{gray!15}62.1 \tiny{(\textcolor{red}{$-6.8$})} \\
\bottomrule
\end{tabular}
}
\vspace{-1.0em}
\end{table}

\begin{table}[h]
  \centering
  \caption{Ablation results on RealText-V1. Comparisons include DocShield (``Reason First''), the FCLE baseline (``Classify First, Explain Later''), uni-modal variants, and CCT component ablations. \textbf{M-F1} is macro-average F1. Best scores are in \textbf{bold}; hyphens (-) indicate metrics not applicable.}
  \label{tab:cct_ablation}
  \resizebox{\columnwidth}{!}{%
    \begin{tabular}{@{}l|cc|cc|cc|c@{}}
      \toprule
      \multirow{2}{*}{\textbf{Configuration}} & \multicolumn{2}{c|}{\textbf{Detection}} & \multicolumn{2}{c|}{\textbf{Grounding}} & \multicolumn{2}{c|}{\textbf{Explanation}} & \multirow{2}{*}{\shortstack{\textbf{M-F1}}} \\
      \cmidrule(l){2-3} \cmidrule(l){4-5} \cmidrule(l){6-7}
       & Acc & F1 & mIOU & mF1 & CSS & BS(F1) & \\
      \midrule
      \rowcolor{gray!15} \textbf{DocShield (Full CCT)} & \textbf{91.4} & \textbf{93.2} & \textbf{32.4} & \textbf{36.7} & \textbf{85.0} & \textbf{76.9} & \textbf{68.9} \\
      \midrule
      FCLE (Full: Vis + Log) & 60.0 & 56.1 & 11.2 & 21.0 & 73.2 & 67.2 & 48.1 \\
      FCLE (Logical Only) & 43.1 & 46.5 & 5.1 & 9.7 & - & - & - \\
      FCLE (Visual Only) & 40.6 & 11.8 & 0.9 & 2.1 & - & - & - \\
      \midrule
      \textit{w/o} Visual Cues & 87.3 & 88.7 & 25.3 & 28.1 & - & - & - \\
      \textit{w/o} Logical Cues & 82.6 & 81.2 & 21.6 & 27.8 & - & - & - \\
      \textit{w/o} Cross-Cues Valid. & 88.7 & 91.6 & 26.7 & 30.2 & 81.3 & 72.5 & 64.3 \\
      \bottomrule
    \end{tabular}%
  }
\end{table}

\noindent\textbf{Robustness Analysis.}
To evaluate DocShield's robustness under real-world conditions, we test it against various visual distortions (Table~\ref{tab:robustness_single}). The model maintains stable performance across distortion types. Under severe Gaussian blur ($K=15$) and JPEG compression ($Q=20$), M-F1 scores remain at 64.2\% and 61.3\%, respectively, exceeding the undistorted baselines of FakeShield and Gemini-2.5-Pro.  
The advantage of visual–logical co-reasoning is evident under OCR-sensitive perturbations. With heavy Gaussian noise ($\sigma=0.3$), Gemini-2.5-Pro's Detection F1 drops by 22.0\%, while DocShield only decreases by 8.5\%. By cross-validating logical cues against degraded visual inputs, the CCT mechanism mitigates localized corruption, ensuring more stable text-centric forgery analysis.

\noindent\textbf{Limitations and Failure Cases.}  
Despite strong performance in most scenarios, DocShield has boundary cases (Appendix~\textbf{E}): (i) false positives in low-resolution or visually chaotic layouts, where benign variations are misinterpreted as manipulations, and (ii) false negatives in flawless, semantically consistent forgeries that evade detection. These cases highlight limitations of cross-cues reasoning under degraded imagery or when manipulations align with internal logical priors, guiding future improvements for more robust document safety.

\section{Conclusion}
\label{sec:conclusion}

In this work, we addressed the challenging task of explainable text-centric forgery analysis to advance AI-based document safety. We introduced DocShield, a unified framework that reformulates the traditionally separate tasks of forgery detection, spatial grounding, and rationale explanation into a joint generative process. This framework is driven by a Cross-Cues-aware Chain of Thought (CCT) mechanism and optimized via a specialized, GRPO-based multi-task reward function. To support this paradigm, we constructed the RealText-V1 benchmark, featuring fine-grained visual and logical annotations generated through a self-correcting PR² multi-agent pipeline.
Extensive experiments validate our approach. DocShield establishes a new state-of-the-art on RealText-V1 while demonstrating strong zero-shot generalization and robustness under extreme structural perturbations on public datasets. These results confirm that logical semantic cues are critical for text-centric forgery analysis, and that visual-logical co-reasoning effectively mitigates the fragile over-reliance on pristine visual inputs. 
Finally, the structured forensic reports generated by DocShield highlight its practical utility. This unified output facilitates downstream applications ranging from automated forensic auditing to providing actionable feedback for refining generative models, ultimately fostering a closed-loop advancement in AI document safety.

\bibliographystyle{ACM-Reference-Format}
\bibliography{main}


\end{document}